\documentclass{article}
\usepackage{spconf,amsmath,graphicx}
\usepackage{hyperref}
\usepackage{url}
\usepackage{booktabs}
\usepackage{amsfonts}
\usepackage{nicefrac}
\usepackage{microtype}
\usepackage{fancyhdr}
\usepackage{algorithmic}
\usepackage{multirow}
\usepackage{float}
\usepackage[capitalize]{cleveref}
\crefname{section}{Sec.}{Secs.}
\Crefname{section}{Section}{Sections}
\crefname{table}{Tab.}{Tabs.}
\usepackage{pifont}

\usepackage{textcomp}
\usepackage{relsize}
\usepackage{xspace}
\usepackage{comment}
\usepackage{amssymb}
\usepackage{bm}
\usepackage{dsfont}

\usepackage{bbm}
\usepackage{threeparttable}
\usepackage{arydshln}
\usepackage{subcaption}
\usepackage{wrapfig}
\usepackage{caption}
\usepackage{xcolor}

\title{Open-World Semantic Segmentation with Sensitivity Modeling}

\name{Anastasios Romanos Varvarigos and Nikos Giakoumoglou and Tania Stathaki}
\address{Department of Electrical and Electronic Engineering,
         Imperial College London,
         London, UK\\
         \texttt{\{anastasis.varvarigos21, nikos, t.stathaki\}@imperial.ac.uk}
}

\begin{document}


\maketitle


\begin{abstract}
Modern vision systems must operate in \textit{``open-world''} settings, where models must recognize known categories and detect unseen or anomalous content. Conventional semantic segmentation models operate under a \textit{``closed-world''} assumption, often producing overconfident misclassifications on novel content. We address open-world semantic segmentation, the joint task of segmenting known classes while detecting and grouping novel or anomalous content without additional supervision, by extending a dual-decoder baseline with a third, complementary decoder within a unified encoder-decoder design. The first decoder performs closed-set segmentation using Gaussian prototypes for known categories. The second uses contrastive feature learning to isolate unknown regions in embedding space. The third, our key contribution, is a sensitivity decoder that captures fine-grained texture irregularities and activation instabilities indicative of semantic uncertainty, which neither semantic prototypes nor contrastive norms can reliably detect. The three decoders provide genuinely complementary signals: class-level OOD distance in logit space, global feature energy in embedding space, and local activation instability across encoder scales. Experiments on Cityscapes and BDD-Anomaly show that our method improves anomaly segmentation and novel-class discovery while maintaining competitive closed-set accuracy, with gains of +2.4\% AUROC and a 2.5 pp.\ reduction in FPR@95TPR on BDD-Anomaly over the baseline.
\end{abstract}




\section{Introduction}
\label{sec:intro}

Semantic segmentation assigns a semantic label (\textit{e.g.}, road, pedestrian, car) to every pixel of an image. Despite significant progress, most segmentation models operate under a \textit{``closed-world''} assumption: they are trained and evaluated on a fixed, predefined set of classes. This assumption fails in real-world deployments, where models frequently encounter out-of-distribution (OOD) content such as unexpected road debris, novel vehicle types, or context-dependent anomalies. Conventional networks tend to produce overconfident misclassifications~\cite{hendrycks2017iclr}, leading to unsafe behavior in safety-critical applications such as autonomous navigation and robotic scene understanding.

\textit{``Open-world''} semantic segmentation addresses the problem of jointly segmenting known classes and detecting novel or anomalous regions without additional supervision~\cite{chan2021neurips}. It combines two complementary sub-tasks: \textit{anomaly segmentation}, which produces a binary map distinguishing known from unknown pixels, and \textit{novel class discovery}, which performs pixel-wise classification of novel samples into coherent, previously unseen categories. Solving both simultaneously requires a model to be simultaneously discriminative (for known classes) and appropriately uncertain (for unknown content), objectives that are in tension under standard training.

Existing dual-decoder approaches~\cite{sodano2024openworldsemanticsegmentationincluding} address these objectives with two complementary branches: a semantic branch that models known class distributions in logit space, and a contrastive branch that structures embedding norms to separate known and unknown features globally. However, both operate at a relatively coarse level of abstraction: prototype distance and embedding norm are scalar, globally aggregated quantities per pixel. In practice, anomalous content often appears with subtle local cues, \textit{i.e.}, irregular texture patterns, atypical activation frequency content, and inconsistencies at object boundaries, that these global signals are not designed to capture. A partially occluded novel object, for instance, may produce a feature norm within the range of known classes while exhibiting locally inconsistent activation patterns that a spatially sensitive detector would flag. Similarly, road surface anomalies such as oil spills or debris may span only a few pixels, lie far from any class prototype boundary, and yet produce characteristic high-frequency activation signatures at intermediate encoder scales, cues that are washed out by the global pooling and hierarchical upsampling operations that the semantic and contrastive decoders rely on. These observations motivate a third, independent signal specifically tuned to local activation analysis.

In this paper, we propose a three-decoder architecture (\Cref{sec:methodology}) that introduces a third, orthogonal objective: sensitivity modeling. The \textit{semantic decoder} performs closed-set segmentation using Gaussian prototypes. The \textit{contrastive decoder} separates known and unknown pixels in feature space via prototype alignment and objectosphere loss. The \textit{sensitivity decoder} models local, fine-grained uncertainty arising from activation irregularities and texture anomalies that the other two branches cannot detect. All three outputs are fused into a unified segmentation map. We evaluate on BDD-Anomaly~\cite{chan2021cvpr} and Cityscapes~\cite{cordts2016cvpr}, reporting both closed-set mIoU and dedicated open-world metrics (AUROC, AUPR, FPR@95TPR), showing consistent improvements over the baseline~\cite{sodano2024openworldsemanticsegmentationincluding} (\Cref{tab:cityscapes-results,tab:bdd-results,tab:bdd-anomaly-metrics}).

Our \textbf{contributions} are: \textbf{(i)} A sensitivity decoder for fine-grained uncertainty modeling, specifically designed to detect ambiguous boundaries and texture anomalies that semantic and contrastive branches cannot resolve, and conceptually distinct from both prototype-based and norm-based uncertainty. \textbf{(ii)} A fully-convolutional three-decoder architecture with complementary losses, trained end-to-end on a single consumer GPU, achieving improved open-world segmentation. \textbf{(iii)} Evaluation with closed-set and open-world metrics (AUROC, AUPR, FPR@95TPR) demonstrating consistent gains, with ablation studies validating each architectural component.


\section{Related Work}
\label{sec:relatedwork}

\subsection{Anomaly Detection and OOD Recognition}
\label{subsec:anomaly-detection-ood}

Open-world segmentation builds on a long line of anomaly detection work, which addresses the realistic assumption that test-time inputs may contain pixels from classes absent at training time and asks a model to flag such pixels as unknown rather than force them into the closed label set. Early methods rely on softmax confidence thresholding~\cite{hendrycks2017iclr} or energy-based scoring~\cite{liu2020neurips} to flag OOD inputs at the image level. Feature-space methods compare per-pixel embeddings to learned class prototypes, treating distance as an unknownness proxy~\cite{dhamija2018neurips}. These approaches share a common limitation: they operate on globally aggregated statistics per pixel, which can miss locally inconsistent but globally plausible content.

\subsection{Pixel-Level Open-World Segmentation}
\label{subsec:pixel-level-segmentation}

Pixel-level extensions of anomaly detection use synthetic OOD data generation~\cite{chan2021cvpr}, or reconstruction error from generative models. The closest prior work, ContMAV~\cite{sodano2024openworldsemanticsegmentationincluding}, formalizes open-world semantic segmentation with a dual-decoder design: one branch for closed-set prediction and one for contrastive anomaly detection via objectosphere loss~\cite{dhamija2018neurips}. RAML~\cite{dong2022raml} and DMLNet~\cite{cen2021dmlnet} both build on a heavier DeepLabV3+ backbone trained with a multi-stage SGD pipeline, closed-set pretraining followed by pixel-level metric learning, with RAML further fine-tuning on a handful of labeled novel-class examples; their closed-set gains therefore come with a larger backbone, a longer multi-stage schedule, and additional supervision that our single-stage, fully unsupervised, end-to-end training on one consumer GPU does not require. Vision-language models have been applied to open-vocabulary anomaly segmentation~\cite{jeong2023cvpr-wzac}, offering strong zero-shot generalization but relying on billion-parameter backbones incompatible with real-time deployment.

\subsection{Positioning of Our Approach}
\label{subsec:positioning}

Our sensitivity decoder is inspired by TruFor~\cite{Guillaro2023TruFor}, which localizes image tampering through learned local texture sensitivity maps. The underlying intuition is that anomalous content leaves high-frequency traces in intermediate activations even when global statistics appear normal. We adapt this idea to open-world segmentation, training the decoder on proxy unknown labels as a lightweight parallel head. This is conceptually distinct from prototype distance (semantic decoder), embedding norm (contrastive decoder), Bayesian uncertainty, and reconstruction error: it captures local activation instability in a single deterministic forward pass with negligible overhead.


\section{Methodology}
\label{sec:methodology}

\subsection{Architecture}
\label{subsec:architecture}

\begin{figure*}[!t]
    \centering
    \includegraphics[width=0.88\linewidth]{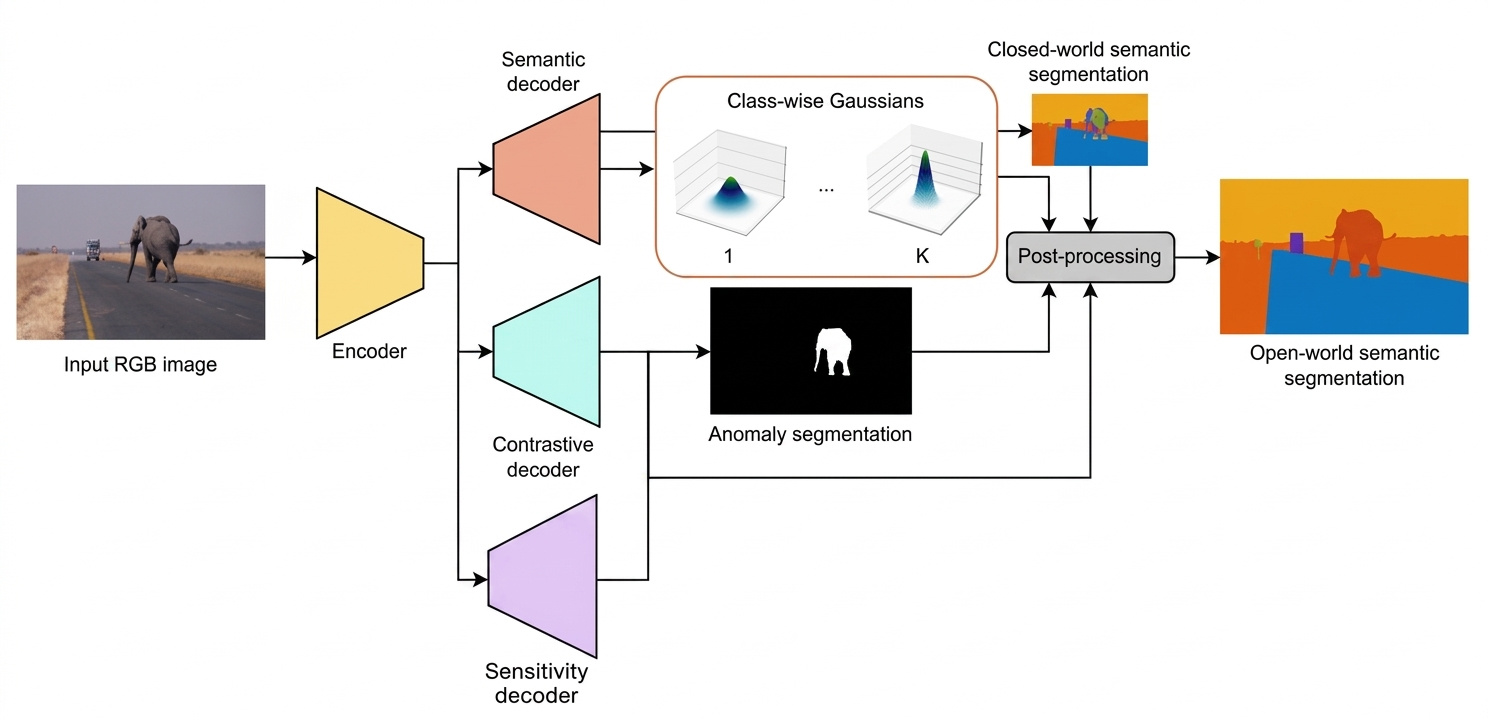}
    \caption{Proposed architecture. A shared ResNet-based encoder with pyramid pooling extracts multi-scale features, which are fed to three parallel decoders: $f_{\text{sem}}$ for semantic segmentation with Gaussian prototype modeling, $f_{\text{con}}$ for contrastive anomaly detection, and $f_{\text{sen}}$ for sensitivity-based uncertainty estimation.}
    \label{fig:our-method}
\end{figure*}

Our architecture uses a shared encoder $f_{\text{enc}}$ and three parallel decoder heads: $f_{\text{sem}}$, $f_{\text{con}}$, and $f_{\text{sen}}$ (\Cref{fig:our-method}). Given input $\mathbf{x} \in \mathbb{R}^{H \times W \times 3}$, the encoder produces a hierarchy of feature maps, of which the deepest is $\mathbf{z} = f_{\text{enc}}(\mathbf{x}) \in \mathbb{R}^{h \times w \times d}$, while intermediate feature maps at multiple resolutions are passed to decoders via skip connections.

\subsection{Semantic Decoder}
\label{subsec:semantic-decoder}

$f_{\text{sem}}$ produces per-pixel logits $\mathbf{y}_{\text{sem}} \in \mathbb{R}^{H \times W \times K}$ for $K$ known classes. We maintain class-wise Gaussian prototypes $\{(\boldsymbol{\mu}_k, \boldsymbol{\sigma}_k^2)\}_{k=1}^K$ computed from correctly classified pixels~\cite{giakoumoglou2025cluster}:
\begin{align}
\boldsymbol{\mu}_k^{(e)} &= \frac{1}{|\hat{\Omega}_k|}\sum_{p\in\hat{\Omega}_k}\mathbf{y}_{\text{sem},p}, \quad \boldsymbol{\sigma}_k^{2,(e)} = \frac{1}{|\hat{\Omega}_k|}\sum_{p\in\hat{\Omega}_k}(\mathbf{y}_{\text{sem},p} - \boldsymbol{\mu}_k^{(e)})^2,
\end{align}
where $\hat{\Omega}_k = \{p \mid \hat{y}_p = y_p = k\}$ is the set of correctly classified pixels for class $k$ at epoch $e$. Prototypes are updated at the end of each epoch using an exponential moving average with momentum $0.9$ for training stability. The feature alignment loss penalizes deviations from class prototypes, normalized by per-class standard deviation for scale invariance:
\begin{equation}
\mathcal{L}_{\text{feat}} = \frac{1}{|\Omega|}\sum_{p\in\Omega}\frac{\|\mathbf{y}_{\text{sem},p} - \boldsymbol{\mu}_{y_p}^{(e-1)}\|_2}{\|\boldsymbol{\sigma}_{y_p}^{(e-1)}\|_2 + \epsilon}.
\end{equation}
The multi-loss semantic objective combines class-weighted cross-entropy $\mathcal{L}_{\text{CE}}$, focal loss $\mathcal{L}_{\text{focal}}$~\cite{lin2017focalloss}, soft dice loss $\mathcal{L}_{\text{dice}}$, and prototype alignment $\mathcal{L}_{\text{feat}}$:
\begin{equation}
\mathcal{L}_{\text{sem}} = \lambda_{\text{CE}}\mathcal{L}_{\text{CE}} + \lambda_{\text{focal}}\mathcal{L}_{\text{focal}} + \lambda_{\text{dice}}\mathcal{L}_{\text{dice}} + \lambda_{\text{feat}}\mathcal{L}_{\text{feat}}.
\end{equation}
This combination is motivated by the severe class imbalance inherent in urban driving datasets: cross-entropy provides stable gradient flow across all classes, focal loss downweights easy background pixels to focus learning on rare foreground and boundary pixels, and dice loss directly optimizes region overlap, complementing pixel-wise losses on small or infrequent objects. In practice, we found that the focal and dice terms together provide sufficient gradient signal for both common and rare classes, and that retaining $\mathcal{L}_{\text{CE}}$ alongside them slightly degraded performance in our final configuration; we therefore set $\lambda_{\text{CE}}{=}0$ in the full model (cf. \Cref{subsec:ablations}, Rows \textcolor{gray}{G}) and \textcolor{gray}{H})), while listing it explicitly in $\mathcal{L}_{\text{sem}}$ for completeness, since intermediate ablation rows do include it.

\subsection{Contrastive Decoder}
\label{subsec:contrastive-decoder}

$f_{\text{con}}$ produces per-pixel embeddings $\mathbf{y}_{\text{con}} \in \mathbb{R}^{H \times W \times d_e}$. The contrastive loss operates on $\ell_2$-normalized embeddings $\bar{\mathbf{y}}_{\text{con}} = \mathbf{y}_{\text{con}}/\|\mathbf{y}_{\text{con}}\|_2$, aligning per-class mean normalized features with historical normalized class prototypes on the unit hypersphere~\cite{giakoumoglou2024review}:
\begin{equation}
\mathcal{L}_{\text{cont}} = -\sum_{k=1}^{K}\log\frac{\exp(\bar{\mathbf{y}}_{\text{con},k}^\top \bar{\boldsymbol{\mu}}_k^{(e-1)} / \tau)}{\sum_{i=1}^{K}\exp(\bar{\mathbf{y}}_{\text{con},k}^\top \bar{\boldsymbol{\mu}}_i^{(e-1)} / \tau)},
\end{equation}
where $\bar{\mathbf{y}}_{\text{con},k}$ is the mean normalized embedding for ground-truth class $k$ pixels and $\tau{=}0.07$ is temperature. The objectosphere loss~\cite{dhamija2018neurips}, by contrast, operates on the raw (unnormalized) embeddings $\mathbf{y}_{\text{con}}$, enforcing radial separation: known pixels ($\Omega_K$) are pushed to have embedding norm $\geq \xi$, while unknown pixels ($\Omega_U$) are pulled toward the origin:
\begin{equation}
\mathcal{L}_{\text{obj}} = \frac{1}{|\Omega|}\!\left[\sum_{p\in\Omega_K}\!\max(0,\xi{-}\|\mathbf{y}_{\text{con},p}\|_2) + \sum_{p\in\Omega_U}\|\mathbf{y}_{\text{con},p}\|_2\right].
\end{equation}
This creates a geometrically interpretable embedding space where the norm of the unnormalized $\mathbf{y}_{\text{con},p}$ directly serves as an unknownness proxy, with the anomaly score at inference given by $a_p^{(c)} = \max(0, 1 - \|\mathbf{y}_{\text{con},p}\|_2/\xi)$. The contrastive objective is $\mathcal{L}_{\text{anom}} = \lambda_{\text{cont}}\mathcal{L}_{\text{cont}} + \lambda_{\text{obj}}\mathcal{L}_{\text{obj}}$.

\subsection{Sensitivity Decoder}
\label{subsec:sensitivity-decoder}

The semantic and contrastive decoders operate on global semantic predictions and feature norms, respectively. However, anomalous regions often manifest locally as subtle texture irregularities, unstable activations near object boundaries, or partially visible novel objects, phenomena that neither semantic prototype distance nor embedding norm can reliably detect. A partially occluded anomalous object may produce a feature norm within the range of known classes while exhibiting spatially inconsistent activation patterns that a local detector would identify. The sensitivity decoder $f_{\text{sen}}$ is specifically designed to capture these fine-grained, local cues.

Inspired by forensic sensitivity networks~\cite{Guillaro2023TruFor}, $f_{\text{sen}}$ processes encoder feature maps from three pyramid levels independently. At each scale, a shared lightweight head comprising three $1{\times}1$ convolutional layers with layer normalization and ReLU activations is applied. The resulting scale-wise sensitivity maps are upsampled to full resolution and fused via element-wise summation, followed by a sigmoid activation to produce the per-pixel uncertainty map $\mathbf{y}_{\text{sen}} \in [0,1]^{H \times W}$. This multi-scale design preserves high-frequency spatial information that the upsampling operations in $f_{\text{sem}}$ and $f_{\text{con}}$ tend to smooth out, making $f_{\text{sen}}$ sensitive to fine-grained local inconsistencies at each resolution level.

Critically, $f_{\text{sen}}$ learns \textit{where the model should be uncertain} rather than \textit{what class a pixel belongs to} or \textit{how far a feature is from known prototypes}. The three decoders therefore provide genuinely complementary signals: semantic prototype distance captures class-level OOD in logit space, contrastive norm captures global feature energy in embedding space, and sensitivity captures local activation instability across encoder scales. The decoder is trained via binary cross-entropy against proxy unknown labels $u_p \in \{0,1\}$, generated from void annotations in the dataset and from OOD patches synthesized via CutPaste augmentation applied to training images:
\begin{equation}
\mathcal{L}_{\text{sens}} = -\frac{1}{|\Omega|}\sum_{p\in\Omega}\!\left[u_p\log \mathbf{y}_{\text{sen},p} + (1{-}u_p)\log(1{-}\mathbf{y}_{\text{sen},p})\right].
\end{equation}

\subsection{Fusion and Inference}
\label{subsec:fusion}

The semantic confidence is derived from the Gaussian likelihood $s_p^{(\text{sem})} = \max_{k}\mathcal{N}(\mathbf{y}_{\text{sem},p} \mid \boldsymbol{\mu}_k, \text{diag}(\boldsymbol{\sigma}_k^2))$, yielding semantic unknownness $s_{\text{unk},p}^{(\text{sem})} = 1 - s_p^{(\text{sem})}$. The contrastive anomaly score and sensitivity map are averaged: $s_{\text{unk},p}^{(\text{cont})} = \frac{1}{2}(a_p^{(c)} + \mathbf{y}_{\text{sen},p})$. The final unknownness score is:
\begin{equation}
s_{\text{unk},p} = \tfrac{1}{2}(s_{\text{unk},p}^{(\text{sem})} + s_{\text{unk},p}^{(\text{cont})}).
\end{equation}
Before fusion, each score is min-max normalized per image to bring all three signals to a common $[0,1]$ range, preventing any single branch from dominating due to scale differences. Pixels with $s_{\text{unk},p} > \delta$ are labeled as unknown; the threshold $\delta{=}0.99$ is selected based on \cite{sodano2024openworldsemanticsegmentationincluding}. Remaining pixels adopt the semantic decoder label $\hat{y}_p$. The full network is trained end-to-end with $\mathcal{L}_{\text{total}} = \mathcal{L}_{\text{sem}} + \mathcal{L}_{\text{anom}} + \mathcal{L}_{\text{sens}}$.


\section{Experiments}
\label{sec:experiments}

\subsection{Implementation Details}
\label{subsec:implementation-details}

The encoder is ResNet-34~\cite{he2016cvpr} with basic residual blocks replaced by NonBottleneck-1D blocks~\cite{romera2018tits} for computational efficiency, augmented with Squeeze-and-Excitation modules~\cite{hu2019squeezeandexcitationnetworks} and a Pyramid Pooling Module~\cite{zhao2017cvpr-pspn}. The semantic and contrastive decoders share a hierarchical architecture of three cascaded modules, each comprising a $3{\times}3$ convolution, NonBottleneck-1D blocks, $2{\times}$ bilinear upsampling, and encoder skip connections via element-wise addition. The sensitivity decoder uses a lightweight multi-scale $1{\times}1$ convolutional MLP applied independently at three encoder scales, with output channels 64, 32, and 16 at the three pyramid levels, before sigmoid-gated element-wise summation. The full model has approximately 28M parameters, of which the sensitivity decoder contributes fewer than 1M ($<$3.6\% of the total), making it a negligible computational addition over the dual-decoder baseline. Inference speed is 24 fps at $1024{\times}512$ resolution on a single NVIDIA L40s GPU, compared to 23 fps for the dual-decoder baseline, confirming the minimal overhead introduced by $f_{\text{sen}}$.

\subsection{Datasets and Training Setup}
\label{subsec:datasets-training}

We evaluate on \textit{Cityscapes}~\cite{cordts2016cvpr} (2,975 train / 500 val, 19 classes, closed-world evaluation) and \textit{BDD-Anomaly}~\cite{chan2021cvpr} (1,935 val images with anomalous objects under diverse weather conditions including rain, night, and fog). Images are resized to $1024{\times}512$ and normalized with ImageNet statistics. We train for 500 epochs using Adam with initial learning rate $4{\times}10^{-3}$, polynomial decay schedule with power 0.9, batch size 8, and weight decay $10^{-4}$. Standard augmentations include random cropping, horizontal flipping, color jitter, and multi-scale resizing. Loss weights are set to $\lambda_{\text{CE}}{=}0$, $\lambda_{\text{focal}}{=}1.0$, $\lambda_{\text{dice}}{=}1.0$, $\lambda_{\text{feat}}{=}0.5$, $\lambda_{\text{cont}}{=}1.0$, $\lambda_{\text{obj}}{=}0.5$ in the final model; the objectosphere radius is $\xi{=}1.0$ and the contrastive temperature is $\tau{=}0.07$.

\subsection{Main Results}
\label{subsec:main-results}

\begin{figure*}
    \centering
    \includegraphics[width=0.98\linewidth]{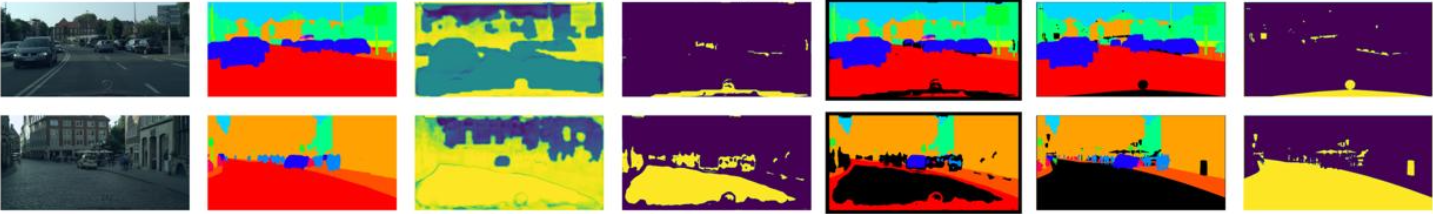}
    \caption{Qualitative results on Cityscapes. \textbf{Columns}: input image, semantic prediction, anomaly logits, anomaly prediction, open-world prediction, semantic GT, anomaly GT. \textbf{Rows}: Our method (bottom) vs. ContMAV \cite{sodano2024openworldsemanticsegmentationincluding} (repr.) (top).}
    \label{fig:cityscapes}
\end{figure*}

\begin{figure*}
    \centering
    \includegraphics[width=0.98\linewidth]{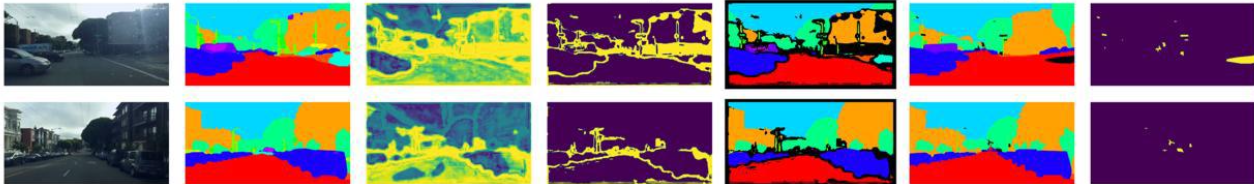}
    \caption{Qualitative results on BDD-Anomaly. \textbf{Columns} and \textbf{rows} are as in \Cref{fig:cityscapes}.}
    \label{fig:bdd}
\end{figure*}

Results on Cityscapes (\Cref{tab:cityscapes-results}) and BDD-Anomaly (\Cref{tab:bdd-results,tab:bdd-anomaly-metrics}) show consistent gains over the reproduced ContMAV baseline across all metrics.\footnote{``--'' indicates metrics not reported by the original method. Entries marked (repr.) are our reproductions under a unified training setup; the gap to the original ContMAV reflects differences in schedule, augmentation, and hardware.} On Cityscapes, the sensitivity decoder and multi-loss landscape provide regularization that benefits closed-set segmentation without degrading it. On BDD-Anomaly, the Recall gain confirms fewer missed detections under challenging conditions, while the AUPR improvement under class imbalance confirms that $f_{\text{sen}}$ provides a signal complementary to the contrastive branch. RAML~\cite{dong2022raml} uses a substantially heavier backbone and multi-stage pipeline; its number is reported for reference only. Qualitative comparisons are shown in \Cref{fig:cityscapes,fig:bdd}.

\begin{table}[!t]
\centering
\caption{Closed-set performance on the Cityscapes validation set. mIoU (\%), Recall (\%), Precision (\%).}
\begin{tabular}{lccc}
\toprule
Method & mIoU & Recall & Precision \\
\midrule
\multicolumn{4}{l}{\textit{Closed-World Methods}} \\
FCN-8s~\cite{long2015fullyconvolutionalnetworkssemantic} & 65.30 & -- & -- \\
DeepLabv2~\cite{chen2017deeplabsemanticimagesegmentation} & 70.40 & -- & -- \\
\midrule
\multicolumn{4}{l}{\textit{Open-World Methods}} \\
DMLNet~\cite{cen2021dmlnet} & 69.20 & -- & -- \\
RAML~\cite{dong2022raml} & 76.70 & -- & -- \\
ContMAV~\cite{sodano2024openworldsemanticsegmentationincluding} & 71.10 & -- & -- \\
ContMAV (repr.) & 69.28 & 89.6 & 69.7 \\
\textbf{Ours} & \textbf{69.88} & \textbf{90.2} & \textbf{70.4} \\
\bottomrule
\end{tabular}
\label{tab:cityscapes-results}
\end{table}

\begin{table}[!t]
\centering
\caption{Closed-set performance on BDD-Anomaly. mIoU (\%), Recall (\%), Precision (\%).}
\begin{tabular}{lccc}
\toprule
Method & mIoU & Recall & Precision \\
\midrule
ContMAV~\cite{sodano2024openworldsemanticsegmentationincluding} & 62.28 & 82.5 & 63.1 \\
ContMAV (repr.) & 56.12 & 78.9 & 60.4 \\
\textbf{Ours} & \textbf{56.68} & \textbf{81.3} & \textbf{62.0} \\
\bottomrule
\end{tabular}
\label{tab:bdd-results}
\end{table}

\begin{table}[!t]
\centering
\caption{Open-world anomaly detection on BDD-Anomaly. AUROC (\%), AUPR (\%), FPR@95TPR (\%; lower is better).}
\begin{tabular}{lccc}
\toprule
Method & AUROC & AUPR & FPR@95TPR $\downarrow$ \\
\midrule
ContMAV (repr.) & 63.4 & 41.8 & 67.4 \\
\textbf{Ours} & \textbf{65.8} & \textbf{44.3} & \textbf{64.9} \\
\bottomrule
\end{tabular}
\label{tab:bdd-anomaly-metrics}
\end{table}

\subsection{Ablation Study}
\label{subsec:ablations}

\Cref{tab:cityscapes-ablation} presents an incremental ablation on Cityscapes. Gaussian prototype alignment $\mathcal{L}_{\text{feat}}$ (Row \textcolor{gray}{C}) is the most impactful component, providing a +4\% gain: normalizing by per-class standard deviation is key, as otherwise the loss is dominated by high-variance classes and regularizes compact categories more weakly. Row \textcolor{gray}{D} recovers the reproduced ContMAV baseline, validating our implementation. The sensitivity decoder (Row \textcolor{gray}{F}) further improves over the dual-decoder setup (Row \textcolor{gray}{E}), with +0.5\% AUROC and 1.2 pp.\ lower FPR@95TPR on BDD-Anomaly, confirming that $f_{\text{sen}}$ mainly benefits unknown-region detection rather than closed-set accuracy. Rows \textcolor{gray}{G} and \textcolor{gray}{H} show that replacing $\mathcal{L}_{\text{CE}}$ with focal + dice yields a small but consistent gain: once focal loss is active, cross-entropy becomes partially redundant and its uniform weighting slightly biases the optimizer toward over-represented classes, so removing it lets focal/dice shape the loss landscape effectively. All contributions are additive: the monotonic improvement from Row \textcolor{gray}{A} to Row \textcolor{gray}{H} confirms that each component addresses a distinct aspect of the open-world problem.

\begin{table}[!t]
\centering
\setlength{\tabcolsep}{1.5mm}
\caption{Ablation on Cityscapes validation set. Configuration H (full model) achieves the highest performance.}
\label{tab:cityscapes-ablation}
\small
\begin{tabular}{ccccccccc}
\toprule
 & $\mathcal{L}_{\text{CE}}$ & $\mathcal{L}_{\text{obj}}$ & $\mathcal{L}_{\text{feat}}$ & $\mathcal{L}_{\text{cont}}$ & $\mathcal{L}_{\text{dice}}$ & $\mathcal{L}_{\text{sens}}$ & $\mathcal{L}_{\text{focal}}$ & mIoU \\
\midrule
\textcolor{gray}{A} & \checkmark & & & & & & & 63.21 \\
\textcolor{gray}{B} & \checkmark & \checkmark & & & & & & 64.21 \\
\textcolor{gray}{C} & \checkmark & \checkmark & \checkmark & & & & & 68.12 \\
\textcolor{gray}{D} & \checkmark & \checkmark & \checkmark & \checkmark & & & & 69.28 \\
\textcolor{gray}{E} & \checkmark & \checkmark & \checkmark & \checkmark & \checkmark & & & 69.40 \\
\textcolor{gray}{F} & \checkmark & \checkmark & \checkmark & \checkmark & \checkmark & \checkmark & & 69.63 \\
\textcolor{gray}{G} & & \checkmark & \checkmark & \checkmark & \checkmark & & \checkmark & 69.70 \\
\textcolor{gray}{H} & & \checkmark & \checkmark & \checkmark & \checkmark & \checkmark & \checkmark & \textbf{69.88} \\
\bottomrule
\end{tabular}
\end{table}


\section{Conclusion}
\label{sec:conclusion}

We proposed a three-decoder architecture for open-world semantic segmentation, the joint task of segmenting known classes while detecting and grouping novel or anomalous content without additional supervision, extending the dual-decoder ContMAV baseline with a sensitivity decoder that provides a complementary uncertainty signal alongside prototype-based OOD distance and contrastive embedding norm. The sensitivity decoder is deterministic, lightweight ($<$1M parameters), and trained on proxy labels from standard annotations. Experiments on Cityscapes and BDD-Anomaly confirm consistent gains in both closed-set mIoU and open-world metrics (AUROC, AUPR, FPR@95TPR), with ablations validating the additive contribution of each component. Future work will explore learned fusion strategies and richer OOD synthesis for the sensitivity branch.


\bibliographystyle{IEEEbib}
\bibliography{refs}

\end{document}